%% file: main.tex
\documentclass[sigconf, screen]{acmart}
\AtBeginDocument{%
  }
\usepackage{amsmath,amsfonts}
\usepackage{algorithmic}
\usepackage{graphicx}
\usepackage{stfloats}
\usepackage{textcomp}
\usepackage{multirow}
\usepackage{multicol}
\usepackage{booktabs}
\usepackage{hhline}
\usepackage{bbding}
\usepackage[linesnumbered,ruled,vlined]{algorithm2e}
\usepackage{colortbl}
\usepackage{booktabs}
\usepackage{array}
\usepackage{enumitem}
\usepackage[misc]{ifsym}
\usepackage[table,xcdraw]{xcolor}
\usepackage{subfigure}
\usepackage{tikz}
\newlength\szg     \newcommand\hquan[1]{%
\settoheight\szg{#1}%
\tikz[baseline]{
\pgfmathparse{1}
\let\hfs\pgfmathresult
\filldraw (0,\szg/2) circle (\szg/2+0.35ex);
\node[white] at (0,\szg/2) {\makebox[0em][c]{\scalebox{\hfs}[1]{\textbf{#1}}}};
}}
\definecolor{MintGreen}{HTML}{32CD32}
\definecolor{SalmonRed}{HTML}{F08080}
\definecolor{LightGray}{HTML}{DCDCDC}

\setcopyright{none} 
\copyrightyear{2026}
\acmYear{2026}
\acmDOI{XXXXXXX.XXXXXXX}
\acmConference[MICRO 2026]{The 58th IEEE/ACM International Symposium on Microarchitecture}{October 31--November 04, 2026}{Athens, Greece}
\acmISBN{978-X-XXXX-XXXX-X/XX/XX}

\begin{document}



\title{A Motion-Aware Vector Quantization Framework with Centroid Reuse for Efficient VLA Inference}



\author{Zhuoran Song$^1$, Haozhe Jiang$^1$, Chunyu Qi$^1$, Minnan Pei$^2$, Gang Li$^2$, Xiaoyao Liang$^1$, Haibing Guan$^1$*}
\affiliation{
    \department{$^1$School of Computer Science, Shanghai Jiao Tong University, $^2$ Institute of Automation, Chinese Academy of Sciences}
    \country{China}
}
\thanks{Haibing Guan is the corresponding author.}
\email{songzhuoran@sjtu.edu.cn}

\begin{abstract}

Vision-Language-Action (VLA) models have demonstrated strong potential for embodied AI, yet their high inference latency on GPUs limits real-time deployment. Existing accelerators, such as Dadu-Corki, improve efficiency but treat VLA models as full-precision workloads, leaving substantial redundancy in both memory and computation underexploited.

In this paper, we propose VQVLA, an algorithm–hardware co-design framework that accelerates VLA inference by exploiting weight similarity and execution dynamics. We first introduce MotionVQ, a motion-aware vector quantization scheme that dynamically adjusts quantization precision based on the robot's execution state, reducing memory access while preserving task success rate. We then propose a merged-centroid vectorized GEMM paradigm that operates on the codebook–index representation, eliminating redundant multiplications through spatial aggregation and temporal reuse of centroids. To realize these optimizations, we design an accelerator that efficiently supports dynamic precision selection and centroid-reuse computation. Experimental results show that VQVLA achieves 6.5$\times$, 2.8$\times$, $1.9\times$, $3.3\times$, and $4.3\times$ speedup over the A100 GPU, Dadu-Corki, LUT-DLA, CodeGEMM, and ShiftAddLLM, respectively, with negligible accuracy degradation.

\end{abstract}


\maketitle

\input{Introduction}

\input{Background}

\input{Motivation}

\input{Algorithm}
\input{Architecture}

\input{Experiment}

\input{Related}
\input{Conclusion}

\bibliographystyle{plain}
\bibliography{sample-base}

\end{document}

%% file: Introduction.tex
\section{Introduction}

Vision-Language-Action (VLA) models~\cite{kim2024openvla, brohan2023rt, liu2024rdt, kim2025fine,li2023vision} have demonstrated remarkable capabilities in executing complex tasks such as object manipulation and spatial navigation. Leveraging these advances, the field of embodied artificial intelligence (AI) integrates VLA models to enable robots to perceive, reason, and act within the physical world. As a result, VLA models have attracted sustained interest from both academia and industry, serving as a key enabler toward the development of Artificial General Intelligence (AGI).

Despite their strong capability, deploying VLA models in real-world robotic systems still requires efficient execution, especially under edge-device constraints, limited power budgets, and higher control frequency requirements. These constraints make inference efficiency critical for sustaining responsive control and safe interaction in dynamic environments~\cite{huang2025dadu,wang2025vlatest,black2024pi_0,bjorck2025gr00t}. Therefore, reducing VLA execution latency remains an important problem for practical and real-time embodied AI systems.



Recent accelerator designs, such as Dadu-Corki~\cite{huang2025dadu}, improve the efficiency of VLA through algorithmic and architectural optimizations. By predicting multiple future actions within a single VLA inference, they effectively reduce invocation frequency. However, such approaches treat the VLA model as a full-precision black box and do not exploit intrinsic redundancy within the model computation. As VLA models continue to grow rapidly in size, this limitation becomes increasingly critical, leading to substantial memory consumption and massive multiply-accumulate (MAC) operations that hinder efficient deployment, especially on resource-constrained devices.

Vector quantization (VQ) has emerged as a promising approach to address these challenges by representing groups of weights with shared centroids, thereby reducing model size. Compared to conventional weight quantization methods, such as microscaling (MX) quantization~\cite{ramachandran2025microscopiq,hu2026m2xfp,lee2025mx+,hao2025algorithm} and outlier-aware element-wise quantization~\cite{lee2024owq,xue2024oltron,guo2023olive}, VQ achieves higher accuracy at the same compression ratio~\cite{egiazarian2024extreme,tseng2024quip,van2024gptvq}. However, directly applying VQ to VLA inference remains ineffective because \emph{existing VQ approaches rely on static precision policies, ignoring that different stages of robot execution exhibit non-uniform sensitivity to quantization.}



\begin{figure*}[!t]
\centering
\includegraphics[width=0.9\linewidth]{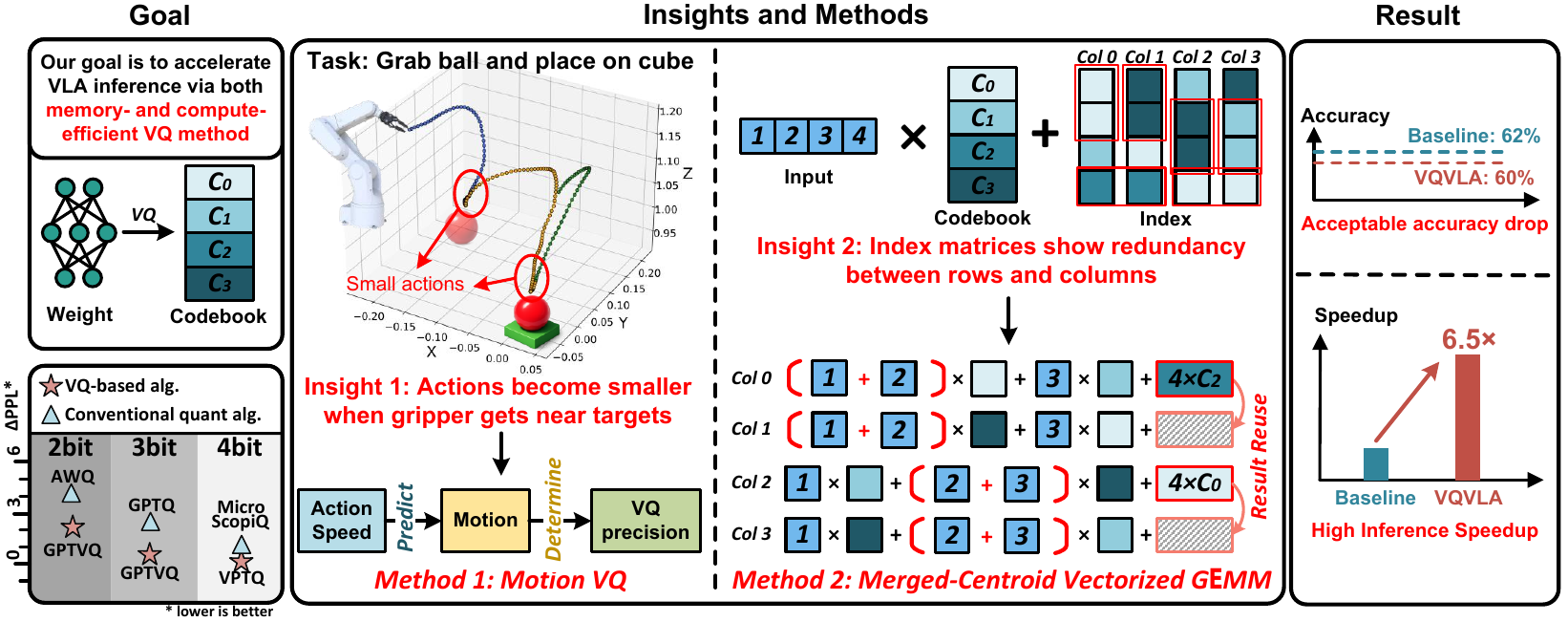}\vspace{-12pt}
\caption{Overview of VQVLA, a hardware and software co-design framework.}\vspace{-12pt}
\label{fig-Introduction_new}
\end{figure*}

To better understand the sensitivity of the robotic system, we analyze the execution behavior of VLA-driven robots and uncover a key insight: \textit{\textbf{the sensitivity of each execution step is strongly correlated with the magnitude of motion between consecutive actions}}. As illustrated in Fig.~\ref{fig-Introduction_new}, when the robot operates near target objects (e.g., grasping or placing the ball), the motion magnitude is small, reflecting fine-grained adjustments required for precise interaction. We refer to this phase as the \textit{execution state}, where the system is highly sensitive to quantization. In contrast, when the robot moves between locations without interacting with objects, the motion magnitude becomes larger, indicating less constrained movements. Such phases are inherently more tolerant to perturbations, and we denote them as the \textit{transition state}. This observation suggests that motion magnitude serves as a natural and lightweight proxy for guiding dynamic VQ.

Building upon this insight, we propose VQVLA, a software-hardware co-design framework that improves both memory and computation efficiency for VLA inference. At the algorithm level, we propose a motion-aware vector quantization (MotionVQ) algorithm that dynamically adjusts quantization precision based on motion magnitude. Specifically, MotionVQ prepares two precision sets offline: a high-precision set with more centroids and higher-bit indices, and a low-precision set with fewer centroids and lower-bit indices. At runtime, the system adaptively selects the appropriate precision based on motion magnitude, using high precision for fine-grained motions and low precision for coarse-grained motions. At the computation level, we observe that the index–centroid representation of VQ introduces substantial redundancy in GEMM execution, as multiple weights are mapped to a limited set of centroids. This observation enables a new execution paradigm that departs from conventional dense GEMM. To exploit this opportunity, we propose Merged-Centroid Vectorized GEMM, which eliminates redundant multiplications by leveraging centroid reuse. Specifically, repeated centroids within a column enable input aggregation prior to multiplication (spatial reuse), while repeated centroids across columns allow reuse of intermediate results (temporal reuse), significantly reducing the number of multiply-accumulate operations. To realize these algorithmic and computation-level optimizations in practice, we further design a customized hardware architecture that directly operates on the codebook and indices, enabling efficient support for centroid reuse and dynamic precision selection.

The main contributions of this paper are as follows:

(1) We identify that VLA inference latency is dominated by the transformer backbone and show that VQ not only achieves higher compression ratios but also exposes strong centroid locality, enabling opportunities for both memory and computation savings.

(2) We propose MotionVQ, a motion-aware vector quantization scheme that dynamically adjusts precision to preserve task success rate, and introduce merged-centroid vectorized GEMM to eliminate redundant computations by exploiting centroid locality.

(3) We design a customized accelerator that jointly supports motion-aware quantization and centroid-reuse computation, translating the theoretical performance improvement to real speedup.

%% file: Background.tex
\section{Background and Related Works}\label{sect:background}

\subsection{Basics of VLA Model}

\begin{figure*}[h]
\centering
\includegraphics[width=0.7\linewidth]{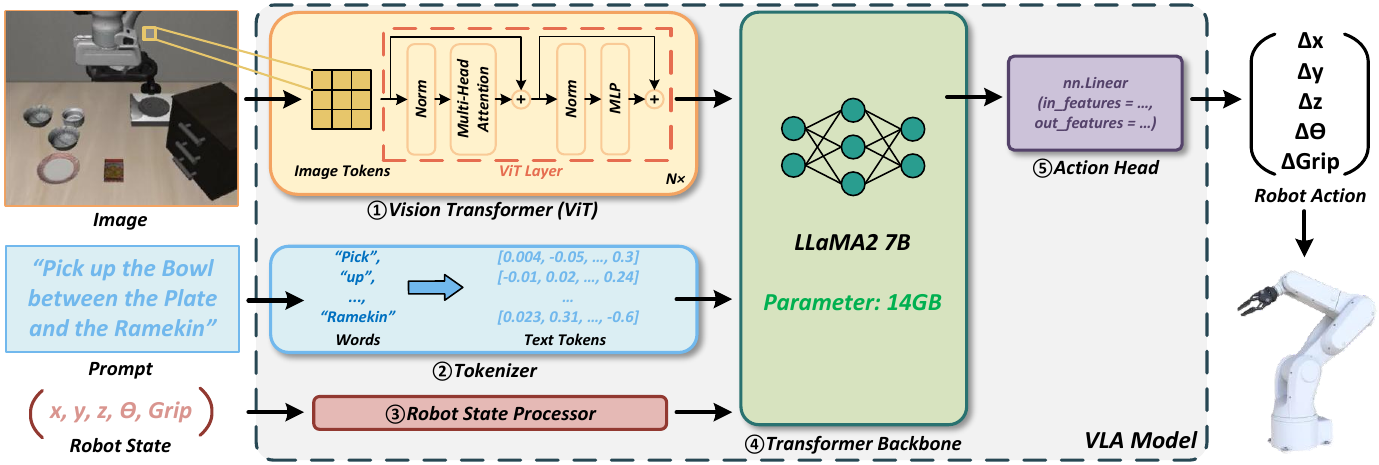}\vspace{-12pt}
\caption{The details of the VLA model.}\vspace{-12pt}
\label{fig-Background-VLA}
\end{figure*}

In this section, we take OpenVLA~\cite{kim2024openvla}, a representative VLA model, as an example to illustrate the basic structure and execution process. The VLA model is primarily designed to control a 6- or 7-degree-of-freedom (DoF) robotic arm. As illustrated in Fig.~\ref{fig-Background-VLA}, the inference of the VLA model mainly consists of three stages. First, the robot sensors capture the current observation, including an image and the robot state, such as joint angles, the gripper's 3D coordinates, and the gripper open/close status. Then, the VLA model takes the sensory observation together with a language instruction, denoted as $Prompt$, as input and predicts the next action. Finally, the robot executes the predicted action. The predicted action is parameterized by five variables: $A_x, A_y, A_z, \Delta \theta$, and $\Delta Grip$. Here, $(A_x, A_y, A_z)$ represent the Cartesian motion magnitudes of the robotic arm, $\Delta \theta$ denotes the gripper rotation, and $\Delta Grip$ indicates the opening and closing states of the gripper.

Fig.\ref{fig-Background-VLA} also shows the structure of the VLA model, which contains four modules: (1) a Vision Transformer (ViT) model, which is responsible for mapping the sensor image to visual tokens; (2) a tokenizer, designed to convert $Prompt$ to text tokens; (3) a transformer backbone (typically using LLaMA2 7B), which packages visual tokens and text tokens as input and extracts feature information from them by conducting GEMMs; (4) an action head, which consists of multiple MLP blocks and decodes the feature representations from the transformer backbone to predict the action. Among these modules, the transformer backbone has substantial parameters and is identified as the primary performance bottleneck targeted in this work.

\subsection{Vector Quantization}

Compared to traditional quantization methods, vector quantization (VQ)~\cite{van2024gptvq,liu2025vq,egiazarian2024extreme,tseng2024quip,zhang2025pqcache} processes multiple weights as a vector rather than quantizing each weight independently. Specifically, VQ treats a weight vector as the basic quantization unit and maps each vector to the index of its nearest centroid in a trained codebook. VQ involves three key configurable parameters: block size, vector size, and the number of centroids in the codebook (denoted as \#Centroid). Specifically, the weight matrix is partitioned into several $m\times m$ blocks. Next, assuming the weight vector size of $1\times n$, each block is split into $\frac{m}{n}$ weight groups, with each group containing $m$ weight vectors that are aligned along the same vertical (y) axis. For each weight group, the vectors are clustered into centroids using the k-means clustering algorithm, where each centroid contains $n$ values. Consequently, each weight vector is replaced by the index of its nearest centroid, forming an index matrix of dimension $m\times (\frac{m}{n})$. Overall, VQ represents the original weight matrix using a codebook and an index matrix. Throughout this process, the block size, vector size, and \#Centroid are tunable. We denote a VQ configuration as VQ$[x,y,z]$, where $x$ is the block size, $y$ is the vector size, and $z$ is \#Centroid. In general, increasing \#Centroid improves the representation capability of the codebook but also increases the codebook size and the bit-width required for each index, thereby reducing the compression ratio.

To perform GEMM with VQ-quantized weights, a dequantization step is required before computation. Specifically, the indices stored in the index matrix are used to retrieve the corresponding centroids from the codebook, and the retrieved centroids are concatenated to reconstruct the original weight groups. The reconstructed weights are then used for floating-point GEMM, where each output element is computed through an inner product between an input row and a reconstructed weight column.

Although VQ algorithms demonstrate notable improvements in accuracy and compression ratios compared to traditional quantization methods, their direct application to VLA models often yields limited performance gains. The reasons include: 1. Existing VQ algorithms fail to observe and exploit the current state of the robotic system. 2. They primarily alleviate the memory bottleneck but fail to reduce the computational load of VLA models.

\subsection{VLA Accelerator Design}

To enhance the inference efficiency of embodied AI systems, Dadu-Corki~\cite{huang2025dadu}, an algorithmic-hardware co-designed framework has been proposed. At the algorithmic level, Dadu-Corki is motivated by the observation that each VLA inference generates a single action. To eliminate frequent calls to the VLA models, it leverages historical robot motion trajectories to predict a sequence of future actions in advance. Specifically, it fits a cubic function to the recent motion trajectory to generate multiple continuous actions for the nearest future. To determine the optimal predicted action length, Dadu-Corki introduces an action-length adaptation mechanism, which detects inflection points in the robot arm's trajectory. This identifies inflection points in the robotic arm, ensuring timely early termination of prediction to maintain a high task success rate. At the hardware level, Dadu-Corki accelerates the robot control process by employing a dedicated accelerator that rapidly converts trajectories into control signals. The accelerator also maximizes intermediate data reuse through customized circuits and data pipelines, enabling high degrees of parallelism and throughput. 

While Dadu-Corki reduces the overall latency of embodied AI systems by minimizing both the number of VLA model invocations and the latency of robot control, it does not exploit the inherent redundancies within the VLA models. In contrast, our work focuses on accelerating the Transformer backbone of VLA models to reduce per-inference latency. Note that our approach operates orthogonally to Dadu-Corki, enabling complementary performance gains when combined.

%% file: Motivation.tex
\section{Motivation}\label{sect:motivation}

\subsection{Challenge}


To gain a more comprehensive understanding of the performance bottlenecks in VLA models, we decompose the inference latency into four stages: tokenizer, ViT, transformer backbone, and action head. We evaluate three VLA models, OpenVLA, OpenVLA-OFT, and RDT, on an NVIDIA A100 GPU. As shown in Fig.~\ref{fig-exp-stage-ratio}, on average, these stages account for 2.7\%, 16.6\%, 80.6\%, and 0.1\% of the total execution time, respectively. These results clearly indicate that the transformer backbone constitutes the primary performance bottleneck:

\begin{figure}[h]
\centering
\includegraphics[width=\linewidth]{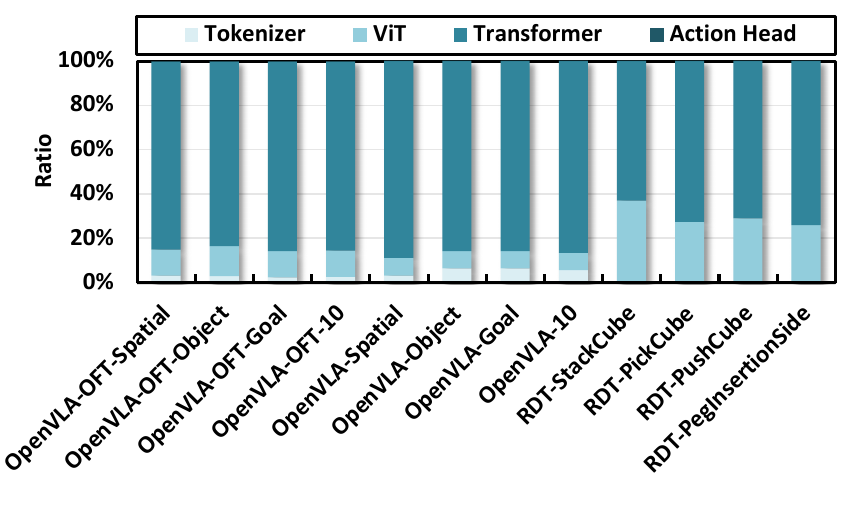}\vspace{-12pt}
\caption{Execution time breakdown of four stages in VLA.}\vspace{-12pt}
\label{fig-exp-stage-ratio}
\end{figure}

1. High Off-Chip Memory Bandwidth Demand: The transformer backbone adopts LLaMA2-7B, which contains 14 GB of parameters. Due to limited on-chip buffer capacity, the transformer requires frequent accesses to off-chip memory, resulting in substantial memory bandwidth pressure. 

2. High Computational Intensity: Each layer of the transformer utilizes eight GEMMs to extract features from environmental information. These include multiple operations with shapes of $N \times 4096 \times 4096$, as well as $N \times N \times 4096$, where $N$ represents the number of input tokens. Since $N$ typically ranges from 200 to 600, these operations result in considerable computational overhead and increased algorithmic complexity.


\subsection{Observation}

\begin{figure}[h]
\centering
\includegraphics[width=0.7\linewidth]{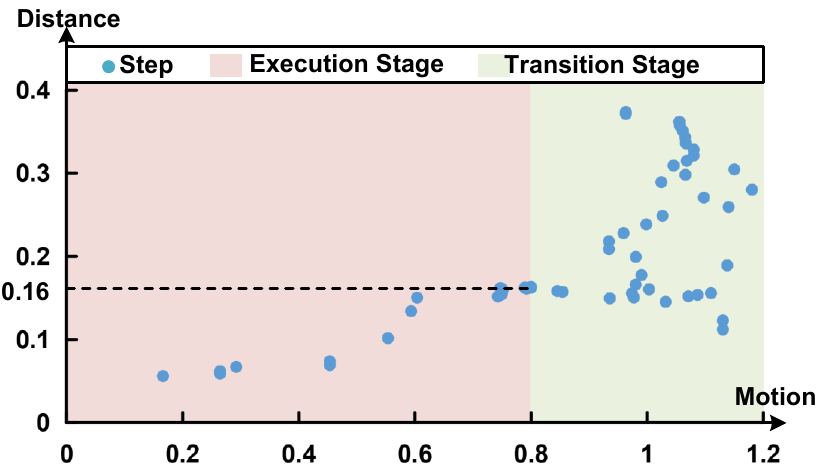}\vspace{-12pt}
\caption{Motion behavior in multi-steps of a task.}\vspace{-12pt}
\label{fig-exp-action-distance}
\end{figure}

In this section, we analyze the motion behavior of the robotic system. As illustrated in Fig.~\ref{fig-exp-action-distance}, the x-axis represents the motion magnitude between consecutive actions, while the y-axis denotes the distance between the robotic arm and nearby objects. Our analysis reveals a clear correlation between motion magnitude and spatial proximity. When the motion magnitude is small (i.e., $x \leq 0.8$), the robotic arm typically operates near target objects ($y \leq 0.16$), where precise adjustments are required. In contrast, larger motion magnitudes (i.e., $x > 0.8$) correspond to phases where the robotic arm is farther from objects ($y > 0.16$), indicating less constrained movements.

Based on the distinction in motion magnitude, we categorize the robot's operation into two states: the execution state and the transition state. During the execution state, the robot performs finer, more controlled movements, which can significantly affect the task success rate. In contrast, the transition state involves coarser motions, where no direct manipulation occurs. We hypothesize that actions taken in this state have minimal impact on the task success rate due to the less precise movements.

\begin{figure}[h]
\centering
\includegraphics[width=0.7\linewidth]{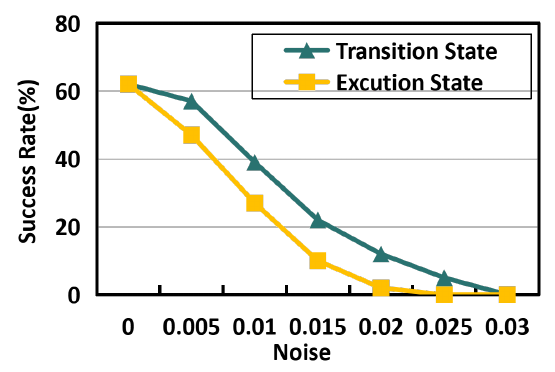}\vspace{-12pt}
\caption{Success rate under different noise patterns.}\vspace{-12pt}
\label{fig-exp-state-motivation}
\end{figure}

To validate the hypothesis regarding the impact of these two states on task success rate, we injected varying levels of noise into both states and evaluated the success rates using the OpenVLA-OFT model on the LIBERO dataset. As shown in Fig.~\ref{fig-exp-state-motivation}, the success rate for the transition state (green line) consistently remained higher than for the execution state (yellow line) under the same noise levels. This demonstrates that the transition state is significantly more tolerant to noise, confirming that it is less sensitive to perturbations compared to the execution state.

\begin{figure}[h]
\centering
\includegraphics[width=0.85\linewidth]{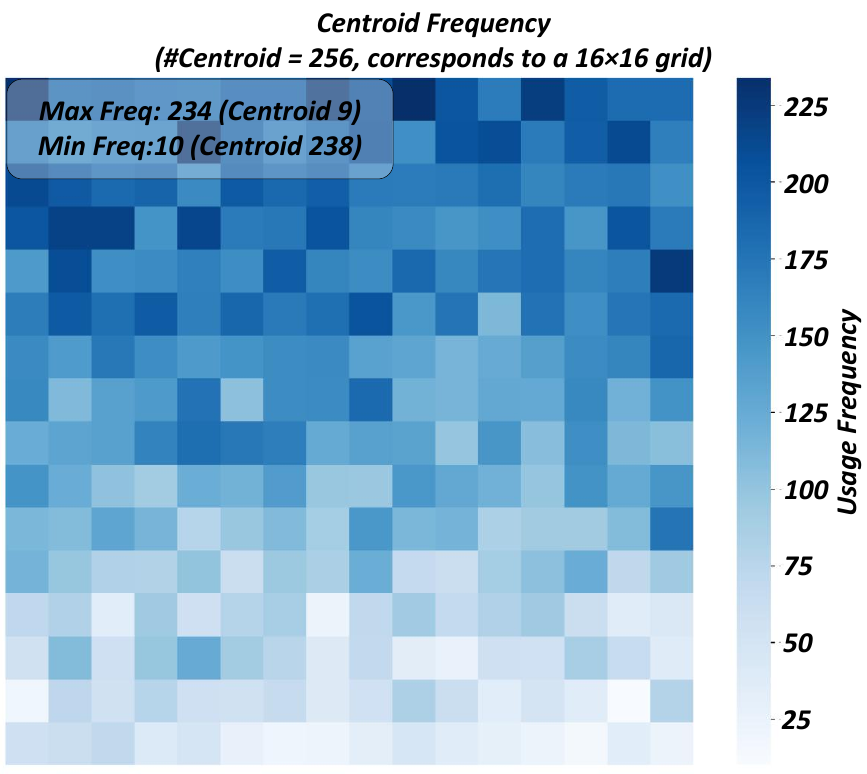}\vspace{-12pt}
\caption{Visualization of centroid access frequency.}\vspace{-12pt}
\label{fig-heatmap}
\end{figure}

To further reduce computational overhead in VLA inference, we analyzed the centroid access frequency after applying VQ. As shown in Fig.~\ref{fig-heatmap}, we visualize the centroid access frequency using a heatmap, where darker colors represent higher access frequencies of centroids. The figure reveals that certain centroids are accessed far more frequently than others, confirming the existence of hot indices. Leveraging this observation, we introduce a merged-centroid vectorized GEMM paradigm to eliminate redundant multiplications for frequently accessed centroids within and across weight matrix columns. 

%% file: Algorithm.tex
\section{MotionVQ Algorithm}\label{sect:algorithm}

Building on the observation that robotic execution exhibits state-dependent sensitivity, we address two key challenges: (1) how to accurately identify the system state with minimal overhead, and (2) how to reduce memory bandwidth consumption without degrading task success rate.

\begin{figure*}[h]
\centering
\includegraphics[width=0.7\linewidth]{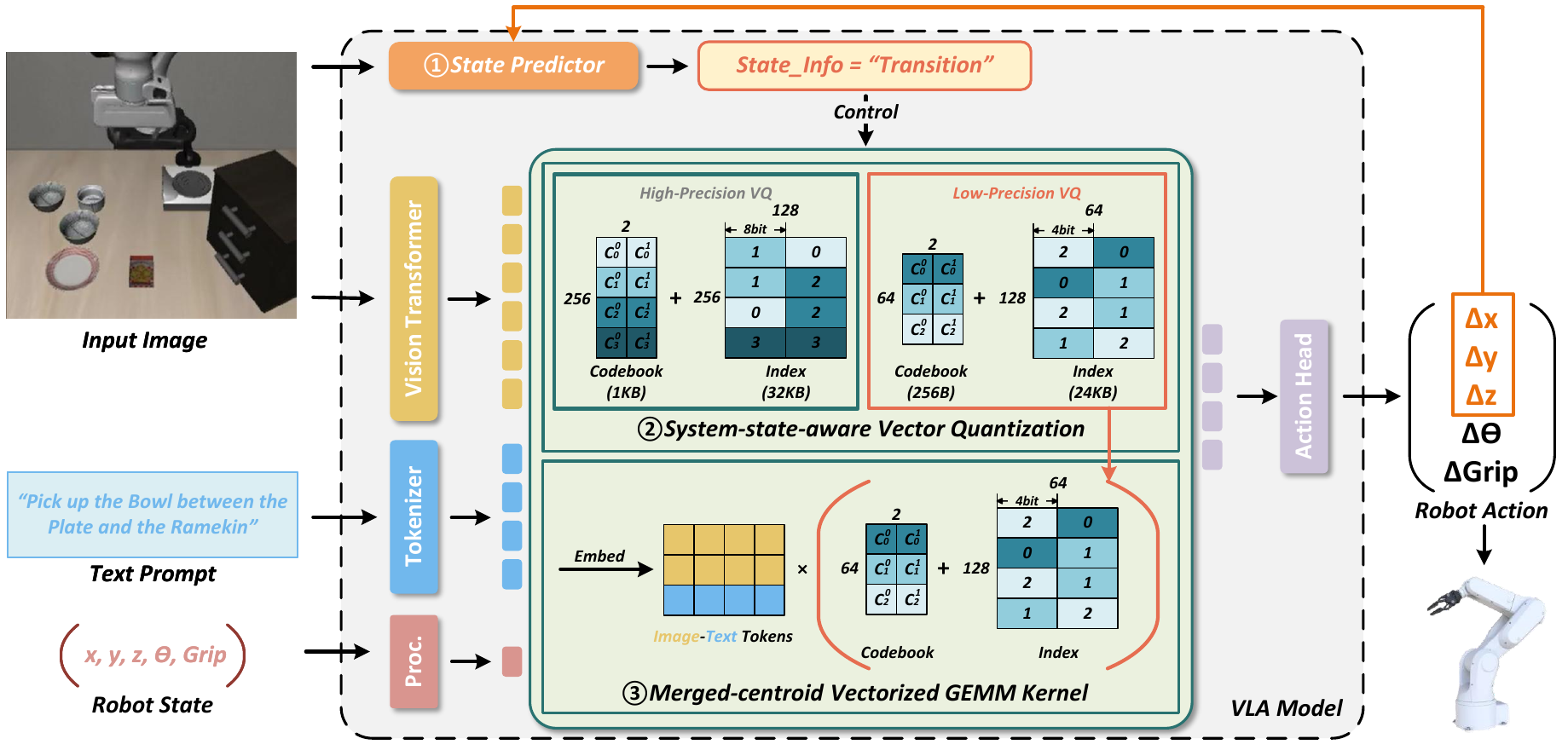}\vspace{-12pt}
\caption{The overview of VQVLA algorithm.}\vspace{-12pt}
\label{fig-Algorithm-Overview}
\end{figure*}

To this end, we propose the \textbf{MotionVQ} algorithm, as illustrated in Fig.~\ref{fig-Algorithm-Overview}. MotionVQ consists of two stages: state prediction and precision adaptation. First, a lightweight state predictor $\hquan{1}$ estimates the current system state based on motion magnitude. Intuitively, larger motion distances indicate transition phases, whereas smaller motions correspond to execution phases. Second, we apply VQ to compress the transformer's weights into two pre-defined precision configurations. Based on the predicted state, the system dynamically selects the corresponding codebook and index matrix $\hquan{2}$.

\textbf{State Prediction.} To determine the current system state, we first compute the motion magnitude $D$ as $D = \sqrt{A_x^2 + A_y^2 + A_z^2}$, where $(A_x, A_y, A_z)$ denotes the 3D action of the robotic arm at each step, obtained from the output of the last VLA inference. The magnitude $D$ is then compared against a predefined threshold $T_d$. When $D \leq T_d$, the robotic arm operates near target objects and performs precise adjustments; thus, the system is classified as the execution state. Otherwise ($D > T_d$), the robot undergoes larger and less constrained movements, corresponding to the transition state.

\textbf{Precision Adaptation.} Based on the predicted state, MotionVQ dynamically selects between two quantization configurations generated offline. The high-precision configuration employs a larger codebook with 256 centroids (\#Centroid = 256), while the low-precision configuration uses a smaller codebook with 64 centroids (\#Centroid = 64). This design enables a state-aware trade-off between memory efficiency and task success rate during VLA inference.

\begin{figure*}[!t]
\centering
\includegraphics[width=0.8\linewidth]{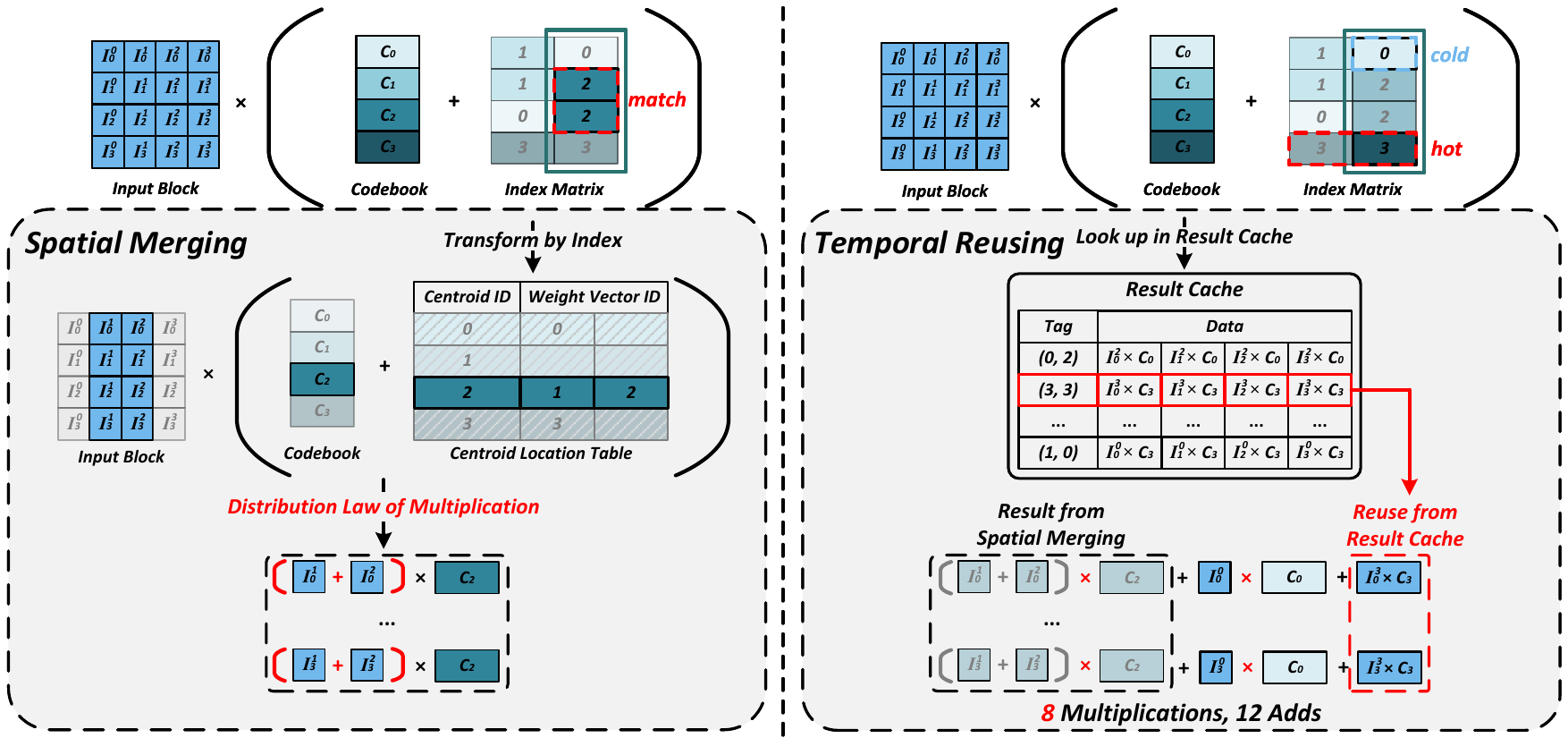}\vspace{-12pt}
\caption{The details of spatial merging (a), temporal reusing (b).}\vspace{-12pt}
\label{fig-Algorithm-Spatial_Temporal}
\end{figure*}

\section{Merged-centroid Vectorized GEMM Paradigm}


In this section, we propose a merged-centroid vectorized GEMM kernel (denoted as $\hquan{3}$ in Fig.~\ref{fig-Algorithm-Overview}) that directly operates on the codebook–index representation to improve the efficiency of VLA inference. By exploiting both spatial and temporal locality of centroids, the kernel avoids redundant computations arising from repeated centroid accesses.

Unlike conventional VQ-based inference, which requires reconstructing floating-point weights followed by standard GEMM, our design eliminates the dequantization step and instead performs computation directly in the compressed domain. This enables efficient reuse of centroids through spatial merging and temporal reusing techniques, significantly reducing computation overhead.

\subsection{Spatial Merging}

Spatial Merging exploits the repetition of centroids within a column (spatial-level hot centroids). In the index matrix, since identical indices indicate that corresponding weight vectors are mapped to the same centroid, multiple multiplications involving the same hot centroid and different input elements can be consolidated into a single operation. This is represented as:
\begin{equation}
    \sum(I_i \times Centroid)\rightarrow \sum(I_i) \times Centroid
\end{equation}

Consider computing the inner product between the $i_{th}$ row of the input matrix $I_i$ and the $j_{th}$ weight group in the weight matrix. When multiple positions $k_1$, $k_2$, ..., $k_s$ in the index matrix share the same index value $idx$ (i.e., $index[k_1, j]$ = $index[k_2, j]$ = ... = $index[k_s, j]$ = $idx$), this indicates that the corresponding input elements $I_i^{k_1}$, $I_i^{k_2}$, ..., $I_i^{k_s}$ must each be multiplied by the same hot centroid ($Centroid=codebook[idx]$). According to the distribution law of multiplication, we can sum $s$ inputs mapped to the same centroid first and then perform a single multiplication with the centroid.


To efficiently implement the spatial merging technique, we construct a specialized data structure called the spatial centroid location table for each weight group. This table is built by traversing the corresponding column of the index matrix. Specifically, when the index value in the $j_{th}$ row is $i$, we record $i$ and $j$ as the centroid ID and weight vector ID in the spatial centroid location table, respectively. This entry indicates that the $j_{th}$ weight vector is mapped to Centroid $i$. Once the spatial centroid location table is constructed, we discard entries associated with only one or zero weight vector IDs, as they do not require input accumulation. For the remaining entries, the corresponding inputs can be accumulated prior to multiplication with their associated centroids. This accumulation eliminates redundant computations and enables efficient spatial merging. As the example in Fig.~\ref{fig-Algorithm-Spatial_Temporal}, when computing the inner product between the input block and the $1_{st}$ weight group, the spatial centroid location table includes an entry for centroid ID "2" and weight vector IDs "1" and "2". This means that the $1_{st}$ and $2_{nd}$ weight vectors within the weight group are clustered into Centroid $2$ (denoted as $C_2$ in the figure). Consequently, taking the $0_{th}$ input row as an example, the spatial merging technique first computes the sum of the inputs ($I_0^1$ and $I_0^2$) that correspond to the $1_{st}$ and $2_{nd}$ weight vectors. Next, it multiplies this sum by $C_2$. This merging procedure is applied to all input rows, so the same accumulation also occurs for $I_1$-$I_3$.


\subsection{Temporal Reusing}

After performing spatial merging, temporal reusing is applied to further reduce the total number of multiplications. Temporal Reusing exploits centroid repetition across columns (temporal-level hot centroids). Its core principle is that multiplications involving identical centroids across columns (and thus identical indices) with the same input can be skipped by reusing cached results of hot centroids.



To realize the temporal reusing technique, we first traverse all indices across columns and apply a Top-$k$ selection offline to determine the most frequently accessed centroids. Note that the offline Top-$k$ centroid selection is not calibrated from environment-specific inputs. Instead, it is determined by the occurrence frequencies of centroids in the compressed weight matrices, which are available before inference. During runtime, we initialize a result cache composed of ($tag$, $data$) entries, where $tag$ is a tuple of the centroid ID and weight vector ID, and $data$ stores the products between activations and these hot centroids. Next, when computing the inner product between the $i_{th}$ row of the input matrix $I_i$ and the $j_{th}$ weight group, redundant multiplications can be eliminated by directly retrieving the matching pre-computed result from the cache if the $tag$ matches. For example, suppose the result cache contains an entry with $tag = (index[k, j-1], k)$ and $data$, where $index[k, j-1]$ denotes the centroid ID and $k$ is the weight vector ID. When processing the $j_{th}$ weight group, the temporal reusing mechanism compares the $j_{th}$ column of the index matrix against cache tags. If a match is found, e.g., $index[k, j-1] = index[k, j]$, the cached $data$ associated with $(index[k, j-1], k)$ can be reused directly.

As manifested in Fig.~\ref{fig-Algorithm-Spatial_Temporal}, given the initialized result cache, which contains a $tag=(3,3)$, and the $1_{st}$ column of the index matrix contains the index "3" ($index[3, 1]=3$ and $k=3$), resulting in a cache hit with $tag=(3,3)$. Therefore, when processing the $1_{st}$ weight group, the cached $data$ including $I_0^3\times C_3$, $I_1^3\times C_3$, $I_2^3\times C_3$, and $I_3^3\times C_3$ can be reused to skip the multiplications for the $3_{rd}$ weight vector of the $1_{st}$ weight group. For mismatched indices, we compute them as usual, as done for the $0_{th}$ weight vector. 

%% file: Architecture.tex
\section{Architecture}\label{sect:architecture}

\subsection{Architecture Overview}

\begin{figure}[h]
\centering
\includegraphics[width=\linewidth]{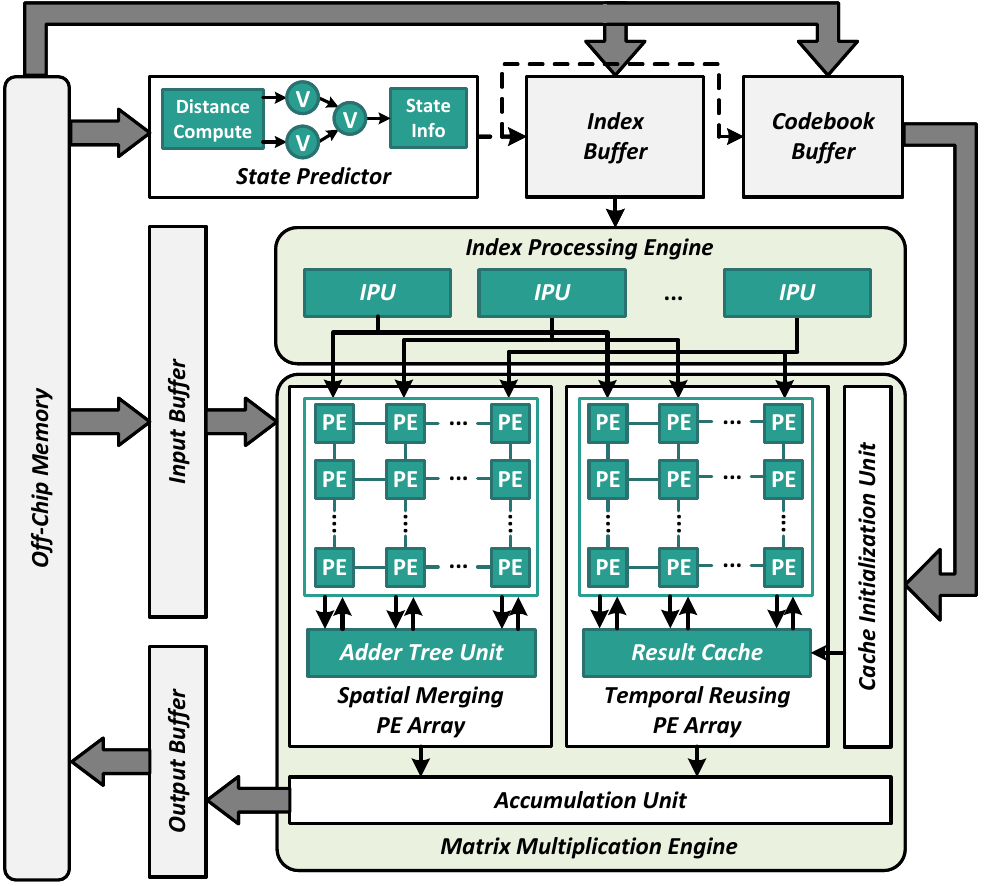}\vspace{-12pt}
\caption{The overview of VQVLA architecture.}\vspace{-12pt}
\label{fig-Architecture-Overview}
\end{figure}

This section presents the VQVLA architecture. As illustrated in Fig.~\ref{fig-Architecture-Overview}, it mainly comprises a state predictor, a matrix multiplication engine, and an index processing engine. The state predictor equips multipliers and adders to predict the current robotic system state. The matrix multiplication engine contains a spatial merging PE array, a temporal reusing PE array, an adder tree unit, a cache initialization unit, a result cache, and an accumulation unit. These components collaboratively realize the spatial merging and temporal reusing mechanisms to enhance the computational efficiency of VLA models. The index processing engine consists of multiple index processing units (IPUs) designed to enable high parallelism in constructing centroid location tables. Each IPU independently generates a spatial centroid location table and a temporal centroid location table, which are then dispatched to the corresponding PEs in the spatial merging array and temporal reusing array, respectively. 

The dataflow of VQVLA can be divided into three steps: 1) The first step is to perform the MotionVQ and fetch the quantized codebook and index matrix. Specifically, the state predictor conducts the motion magnitude calculation and estimates the system state. This predicted state is then used to retrieve the corresponding codebook and index matrix, which are subsequently stored in the codebook buffer and index buffer, respectively. 2) The second step prepares the spatial and temporal centroid location tables with the index processing engine. 3) The third step involves the spatial merging array and adder tree unit to conduct the spatial merging, and the temporal reusing array, cache initialization unit, and result cache to realize the temporal reusing. Since these modules are supported by independent hardware resources, they can be triggered in a pipelined fashion, thereby minimizing the critical path latency.

\subsection{Matrix Multiplication Engine}\label{ssect:matrix-multiplication-engine}

\begin{figure}[!t]
\centering
\includegraphics[width=\linewidth]{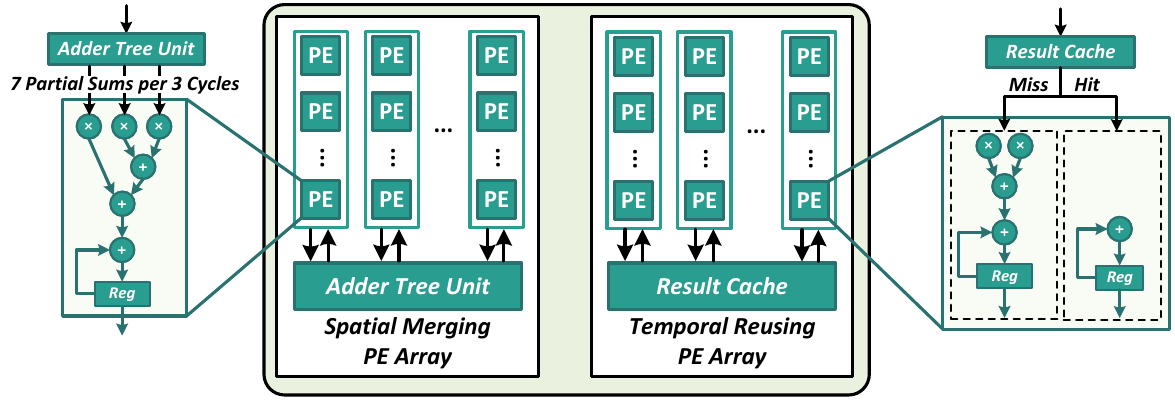}\vspace{-12pt}
\caption{The details of the matrix multiplication engine.}\vspace{-12pt}
\label{fig-Architecture-PE}
\end{figure}

The matrix multiplication engine is the key component of the VQVLA architecture. It implements the spatial merging and temporal reusing mechanisms through two dedicated PE arrays, whose designs are detailed in Fig.~\ref{fig-Architecture-PE}.

\begin{figure*}[h]
\centering
\includegraphics[width=0.7\linewidth]{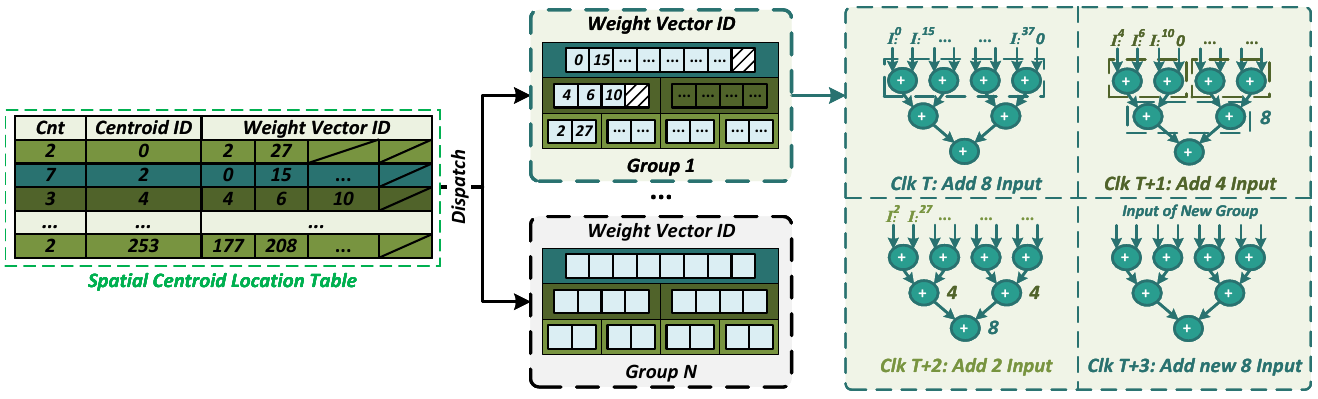}\vspace{-12pt}
\caption{The dataflow of the adder tree.}\vspace{-12pt}
\label{fig-Architecture-Adder_Tree}
\end{figure*}

\subsubsection{Spatial Merging PE Array}

The spatial merging PE array comprises $m\times n$ PEs, which perform output-stationary GEMM operations on $m$ rows of inputs and $n$ columns of weights. To support input accumulation, the array integrates an adder tree unit that accumulates inputs before sending them to the corresponding PEs. This unit includes multiple adder trees, directly connecting to PEs. A critical design challenge for the adder tree arises from the variable number of weight vectors mapped to each centroid (denoted as $Cnt$ in the centroid location table). To balance performance and resource efficiency, we implement a three-stage pipelined adder tree. As evaluated in Section~\ref{ssect:exp-exploration}, this configuration achieves high utilization. 
The adder tree supports three reduction modes over three cycles: one 8-input reduction at Cycle $T$, two parallel 4-input reductions at Cycle $T+1$, and four parallel 2-input reductions at Cycle $T+2$. Thus, a fully packed group can produce seven accumulated results over three cycles by reusing different levels of the same fixed adder tree. As shown in Fig.~\ref{fig-Architecture-Adder_Tree}, each group contains three rows: the first row stores one set of up to eight weight-vector IDs, the second row stores two sets of up to four IDs, and the third row stores four sets of up to two IDs. During execution, the adder tree retrieves the inputs corresponding to each row, e.g., $I_:^0$, $I_:^{15}$, ... for the 8-input row, $I_:^4$, $I_:^6$, $I_:^{10}$, ... for the 4-input rows, and $I_:^2$, $I_:^{27}$, ... for the 2-input rows, and performs the corresponding reductions.

To supply the adder tree with properly sized input groups, the spatial centroid location table is divided into multiple sets, with each set containing at most eight weight-vector IDs. For $Cnt>8$, the entry is split into multiple sets and sequentially fed into the adder tree. For example, if $C1$ has $Cnt=15$, it is decomposed into one 8-element set and one 7-element set, where the latter is padded to eight lanes before dispatch. The partial sums from these sets are then accumulated for the same centroid. For smaller $Cnt$ values, the dispatch logic maps each set to the appropriate row: a set with $4<Cnt\leq8$ is dispatched to the 8-input row, a set with $2<Cnt\leq4$ to the 4-input row, and a set with $0<Cnt\leq2$ to the 2-input row. For instance, if $C2$ has $Cnt=4$, it is directly dispatched to a 4-input reduction row. Once a group is filled with one 8-input set, two 4-input sets, and four 2-input sets, it can be issued to the fixed adder tree and completed within three cycles.

Moreover, since the adder tree produces 7 accumulated results every three cycles, each PE in the array incorporates $\frac{7}{3}\approx 3$ multipliers and associated adders to process these results within the next three cycles, as shown in Fig.~\ref{fig-Architecture-PE}(a). This design enables a pipelined execution between input accumulation and multiplication, sustaining high throughput.

\begin{figure*}[!t]
\centering
\includegraphics[width=0.7\linewidth]{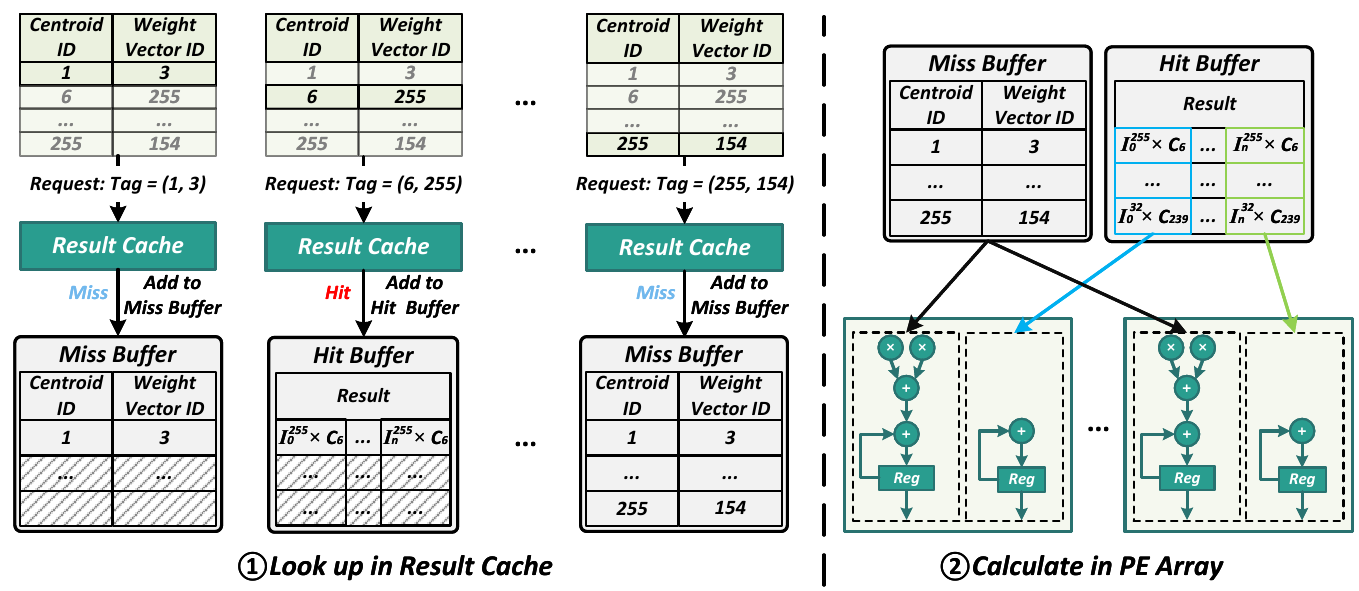}\vspace{-12pt}
\caption{The procedure of the temporal reusing PE array performing temporal reusing.}\vspace{-12pt}
\label{fig-Architecture-Result_Cache}
\end{figure*}

\subsubsection{Temporal Reusing PE Array}

The temporal reusing PE array also comprises $m \times n$ PEs, along with an integrated result cache and a cache initialization unit. Leveraging the offline-identified hot centroids, the initialization unit computes their multiplication results in advance and then stores them in the result cache. Afterwards, the PEs check the result cache to determine whether the current multiplication can be avoided. As detailed in Fig.~\ref{fig-Architecture-Result_Cache}, the temporal reusing PE array's operation can be divided into two steps: 

1) Look up in the result cache: Using the temporal centroid location table, the array generates a request tag comprising a tuple of the centroid ID and weight vector ID. This tag is then used to search the result cache for a match. If a cache hit occurs, indicating that the current multiplication can be bypassed, the previously computed results are retrieved and placed into the hit buffer, awaiting reuse. If a cache miss occurs, meaning that the multiplication is still required, the centroid ID and weight vector ID are sent to the miss buffer, awaiting processing.

2) Calculate in the PE array: After traversing all entries in the temporal centroid location table, the hit and miss buffers trigger the PE array to perform the required computations. Specifically, the hit buffer instructs the PEs to accumulate partial multiplication results using adders. In the meantime, the miss buffer directs the PEs to compute the necessary multiplications and accumulations using both multipliers and adders, with inputs being retrieved based on the centroid ID and weight vector ID. Finally, each PE incorporates an adder to combine the results from the hit and miss paths.

\subsection{Details of State Predictor and Index Processing Engine}

To support efficient execution of the merged-centroid vectorized GEMM in VQVLA, we design two complementary hardware components: a lightweight state predictor and an index processing engine. The state predictor enables runtime state prediction, while the index processing engine constructs spatial and temporal centroid location tables and supplies them to the matrix multiplication engine.

\begin{figure}[h]
\centering
\includegraphics[width=\linewidth]{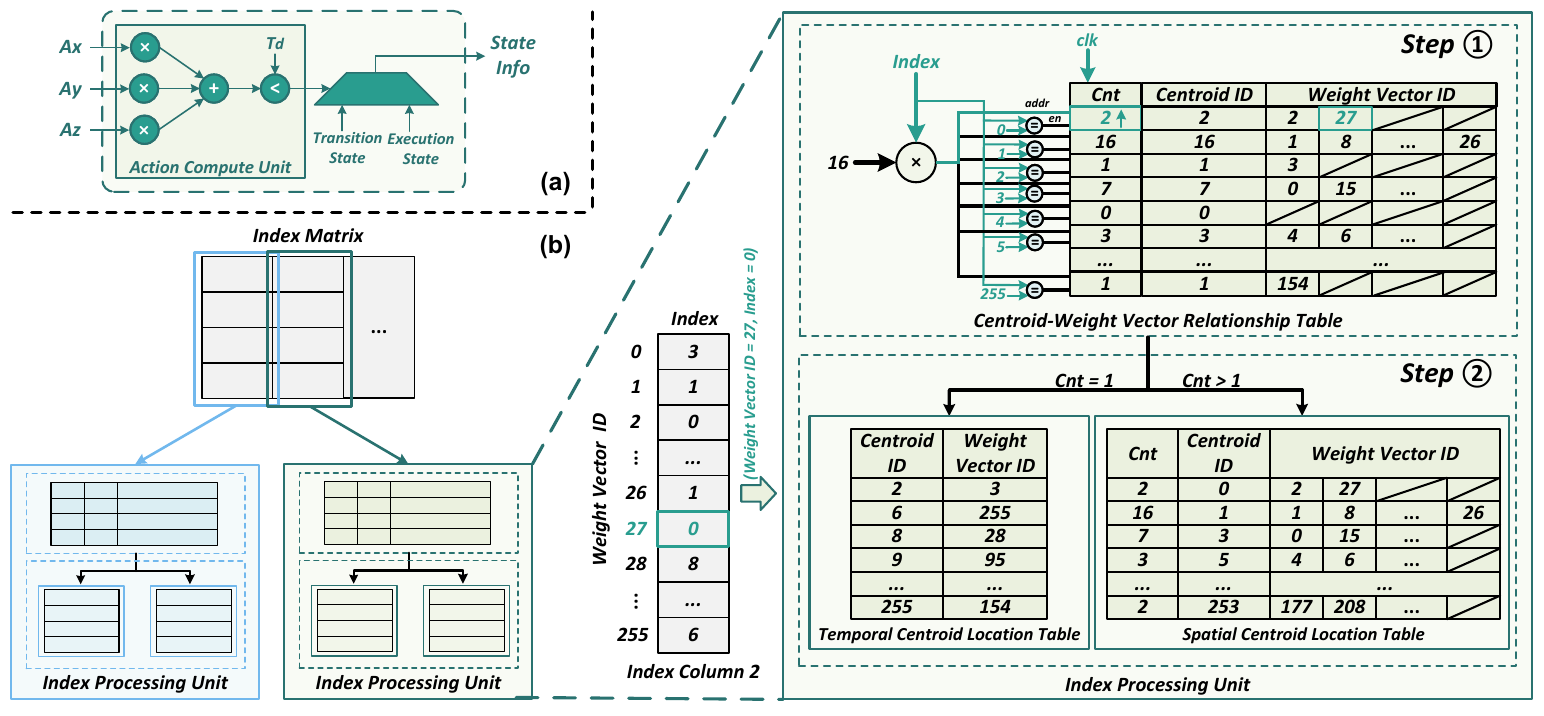}\vspace{-12pt}
\caption{The structure of the state predictor and index processing engine.}\vspace{-12pt}
\label{fig-Architecture-State_Predictor}
\end{figure}

The state predictor, shown in Fig.~\ref{fig-Architecture-State_Predictor}(a), consists of a lightweight motion-magnitude computation unit followed by a threshold comparator. The computation unit evaluates the motion magnitude using three multipliers and an adder, while the comparator determines whether the result is below a predefined threshold $T_d$. If $D \leq T_d$, the system is classified as the execution state; otherwise, it is classified as the transition state.


The index processing engine consists of multiple IPUs, each responsible for generating the spatial and temporal centroid location tables for a given weight group. Each IPU first traverses a column of the index matrix to record which centroid each weight vector is mapped to, producing a centroid-weight-vector relationship table. Taking the example illustrated in Fig.~\ref{fig-Architecture-State_Predictor}(b), if the index value in the $27_{th}$ row is $0$, we treat $0$ as the centroid ID and $27$ as the weight vector ID. The IPU then inserts $27$ into the $0_{th}$ row of the table, indicating that the $27_{th}$ weight vector maps to Centroid $0$. The associated count ($Cnt$) for Centroid $0$ is incremented to 2, showing that two weight vectors are currently mapped to it.

Once the column traversal is complete, the IPU partitions the relationship table into spatial and temporal centroid location tables according to $Cnt$. If $Cnt>1$, meaning multiple weight vectors share the same centroid and require input accumulation, the centroid ID and its associated weight vector IDs are placed into the spatial centroid location table for spatial merging. Otherwise, if $Cnt=1$, the entry is recorded in the temporal centroid location table for temporal reusing.

%% file: Experiment.tex
\section{Evaluation}\label{sect:eval}

\subsection{Workloads}\label{ssect:setup}

To validate the effectiveness of VQVLA across diverse VLA workloads, we evaluate five representative models: OpenVLA~\cite{kim2024openvla}, OpenVLA-OFT~\cite{kim2025fine}, RDT~\cite{liu2024rdt}, $\pi_0$~\cite{black2024pi_0}, and GR00T~\cite{bjorck2025gr00t}. For OpenVLA, OpenVLA-OFT, $\pi_0$, and GR00T, we use the LIBERO simulation benchmark~\cite{liu2023libero}, which models a Franka Emika Panda robotic arm and includes four task suites: LIBERO-Spatial, LIBERO-Object, LIBERO-Goal, and LIBERO-10. For RDT, we use the ManiSkill simulation benchmark~\cite{taomaniskill3}, which covers diverse robot embodiments, including humanoids, mobile manipulators, and single-arm robots, as well as diverse manipulation tasks, including table-top, drawing/cleaning, and dexterous manipulation tasks. We evaluate RDT on four ManiSkill task suites: ManiSkill-PickCube, ManiSkill-PushCube, ManiSkill-PegInsertionSide, and ManiSkill-StackCube. To faithfully reflect the execution flow of each model, we also preserve their default action-chunking settings, where each inference call predicts multiple future actions to reduce the VLA invocation frequency. Specifically, OpenVLA-OFT, RDT, $\pi_0$, and GR00T use action chunk lengths of 8, 8, 5, and 16, respectively.

\subsection{VQVLA Algorithm Evaluation}

\textbf{Methodology.} 
We adopt open-source implementations of the aforementioned VLA models, running on the PyTorch framework \cite{paszke2019pytorch}. We implement the proposed VQVLA algorithm in Python and integrate it into the models' implementations. In our experiments, the high-precision set is configured as VQ$[256, 2, 256]$, and the low-precision set as VQ$[128, 2, 64]$. These configurations correspond to average bitwidths of 4.125 bits and 3.125 bits for the high- and low-precision sets, respectively. We use the task success rate (higher is better) to measure the accuracy of each workload. Success rate refers to the proportion of successful tasks in the total tasks, which can be used to measure the model's ability to complete tasks.

\begin{figure*}[h]
\centering
\includegraphics[width=0.8\linewidth]{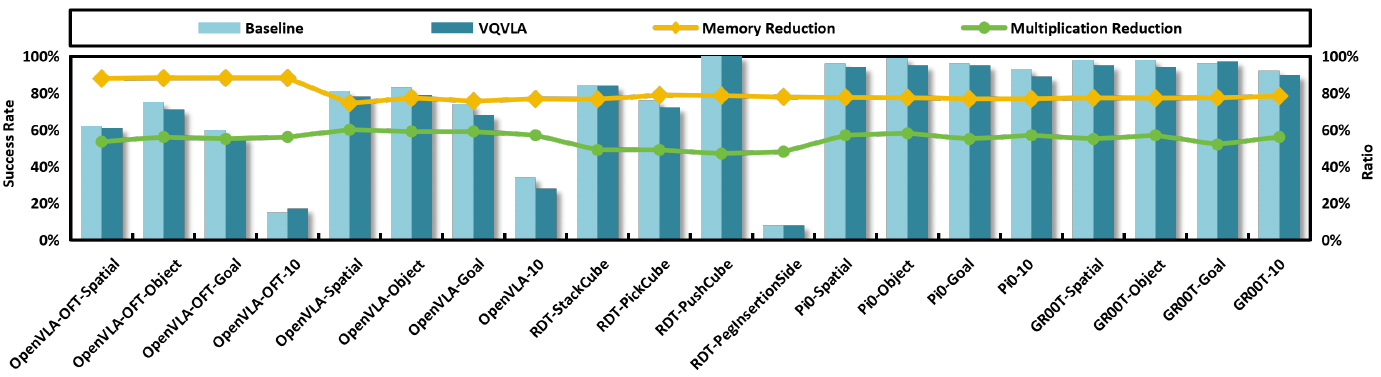}\vspace{-12pt}
\caption{The results of success rate, memory reduction, and multiplication reduction.}\vspace{-12pt}
\label{fig-exp-success-rate}
\end{figure*}


\textbf{Success rate and theoretical performance improvement.} As illustrated in Fig.~\ref{fig-exp-success-rate}, compared to the baseline, the VQVLA algorithm exhibits average reductions in the success rate of $2.5\%$. Such success rate reduction is acceptable, as verified in research papers~\cite{kim2024openvla}. As for the theoretical performance improvement, we first give the reduction of memory access in the figure, where a higher proportion of transition states corresponds to greater memory footprint savings, as the low-precision set is fetched during such states. On average, our proposal reduces the memory consumption for retrieving weights by 79.4\% compared to the baseline model. Additionally, Fig.~\ref{fig-exp-success-rate} presents the reduction of multiplications, where a higher proportion indicates greater efficiency. On average, the merged-centroid vectorized GEMM kernel reduces $54.8\%$ multiplications. It is important to note that the efficiency of our method is closely influenced by task complexity, which results in varying degrees of compute savings across different models and tasks. For example, VQVLA achieves a 71.4\% reduction in multiplications for the OpenVLA model on the LIBERO-Object task, whereas for the RDT model on the ManiSkill-StackCube task, it only reduces multiplications by 52.2\%.

\subsection{VQVLA Architecture Evaluation}\label{ssect:exp-arch}



\textbf{Methodology.} 
To evaluate the performance of the VQVLA architecture, we develop a cycle-level simulator to collect the latency statistics of multiplications, additions, and the number of buffer accesses for each workload. The simulator is integrated with Ramulator~\cite{kim2015ramulator} for off-chip memory timing. We make efforts to ensure the accuracy of the simulator by following the widely adopted open-source simulator, Scale-Sim~\cite{samajdar2018scale}. Moreover, we implement the proposed VQVLA architecture in Verilog and synthesize it using Synopsys Design Compiler to get the chip area and power under 28nm technology with a frequency of 500 MHz. The energy consumption of off-chip memory accesses is estimated at 3.9 pJ/bit~\cite{o2017fine}. Given that the baseline A100 GPU operates on a 7nm node, we scale the area and power measurements of VQVLA to the 7nm node, based on the method described in~\cite{villa2014scaling}.

We compare VQVLA with NVIDIA A100 GPU, two state-of-the-art accelerators, LUT-DLA~\cite{li2025lut} and Dadu-Corki~\cite{huang2025dadu}, and two GPU-based quantization methods, CodeGEMM~\cite{park2026codegemm} and ShiftAddLLM~\cite{you2024shiftaddllm}.

LUT-DLA is a VQ-based accelerator that precomputes activation–weight partial sums and stores them in a lookup table, which are then indexed at runtime. Dadu-Corki is a VLA-specific accelerator that reduces inference frequency by predicting multiple actions per step and overlaps communication with robot execution to hide latency. To compare with Dadu-Corki and LUT-DLA, we re-implement them within our cycle-accurate simulator to ensure consistent modeling and comparability. 

CodeGEMM is a codebook-centric GPU acceleration method for low-bit quantized models. It replaces on-the-fly dequantization with precomputed centroid--activation partial sums, thereby reducing cache pressure and redundant computation. ShiftAddLLM is a post-training reparameterization framework that accelerates inference on GPUs by converting weight multiplications into hardware-friendly shift-and-add operations. We evaluate CodeGEMM and ShiftAddLLM using their official open-source implementations.

For a fair comparison with the GPU baselines, VQVLA adopts the same external memory configuration as A100, using 80GB HBM2e with a peak bandwidth of 1935GB/s. For GPU execution, we use the default precision of each model: FP16 for OpenVLA, FP32 for OpenVLA-OFT, and BF16 for RDT, Pi0, and GR00T. We enable Tensor Core and FlashAttention 2 to optimize the inference speed of VLA models.

\begin{figure*}[h]
\centering
\includegraphics[width=0.8\linewidth]{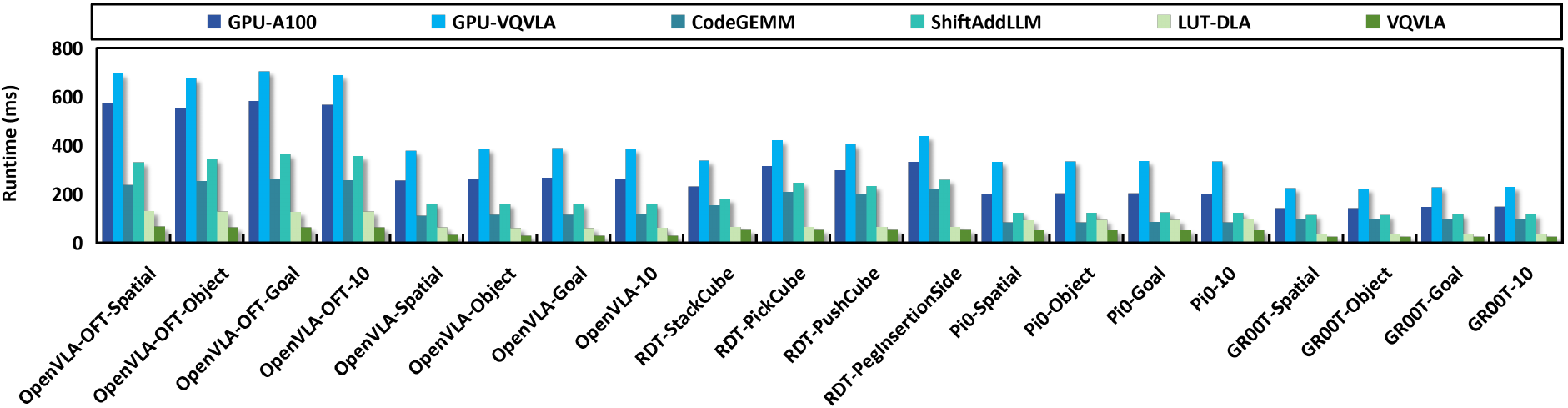}\vspace{-12pt}
\caption{The VLA inference time of GPU, LUT-DLA, CodeGEMM, ShiftAddLLM, and VQVLA architecture.}\vspace{-12pt}
\label{fig-exp-speedup}
\end{figure*}

\textbf{Performance Improvement over GPU, LUT-DLA, CodeGEMM, and ShiftAddLLM.} 
Fig.~\ref{fig-exp-speedup} showcases the inference time of the GPU running the original VLA models (marked as GPU-A100 in the figure), A100 running the VQVLA algorithm (marked as GPU-VQVLA), LUT-DLA, CodeGEMM, ShiftAddLLM, and the VQVLA architecture. On average, the VQVLA architecture achieves $6.5\times$, $1.9\times$, $3.3\times$, and $4.3\times$ speedup over GPU-A100, LUT-DLA, CodeGEMM, and ShiftAddLLM respectively. The performance improvement over GPUs stems from several factors: 1) VQVLA lessens stress on the main memory by adaptively fetching codebooks and indices based on the real-time robotic system state, instead of retrieving all FP32 data. 2) VQVLA reduces the unnecessary multiplications by spatial merging and temporal reusing techniques. 3) The specialized hardware modules, such as the state predictor and matrix multiplication engine, enable highly parallelized operations such as the robotic system state prediction and the merged-centroid vectorized GEMM. In contrast, the GPU executes kernels serially.

\begin{figure}[h]
\centering
\includegraphics[width=0.9\linewidth]{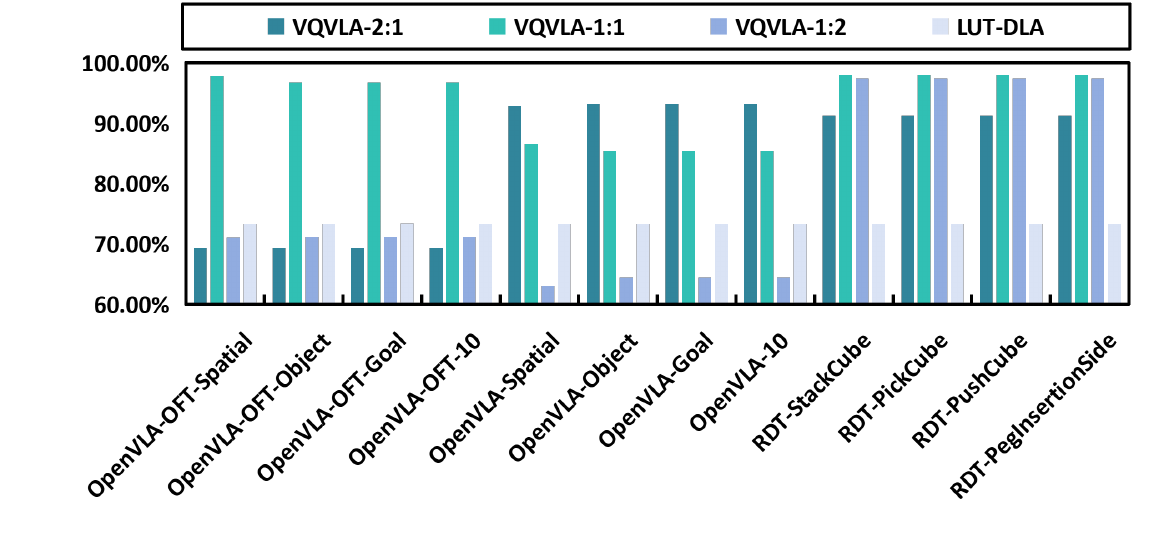}\vspace{-12pt}
\caption{The PE utilization of VQVLA and LUT-DLA.}\vspace{-12pt}
\label{fig-exp-pe-utilization}
\end{figure}

VQVLA outperforms LUT-DLA for two main reasons. First, although LUT-DLA eliminates multiplications via lookup operations, it requires large lookup tables to store precomputed partial sums, resulting in substantial area overhead. This overhead is particularly significant for VLA models, where weights and activations cannot be aggressively quantized to ultra-low precision. In contrast, VQVLA achieves a more area-efficient design, allowing more hardware resources to be allocated to computation. Second, LUT-DLA suffers from irregular table accesses caused by runtime index lookups, which introduce severe on-chip buffer bank conflicts and lead to underutilized compute resources. To further quantify this effect, we report the PE utilization in Fig.~\ref{fig-exp-pe-utilization}. VQVLA (denoted as VQVLA-1:1) achieves 90.6\% PE utilization, higher than the 73.3\% achieved by LUT-DLA, demonstrating that VQVLA more effectively exploits data reuse and sustains compute resource utilization.

As illustrated in Fig.~\ref{fig-exp-speedup}, VQVLA surpasses CodeGEMM and ShiftAddLLM by $3.3\times$ and $4.3\times$, respectively. Although CodeGEMM and ShiftAddLLM reduce multiplications and memory accesses, their fine-grained gather, shift, and add operations are difficult to map efficiently onto Tensor Cores. As a result, they cannot fully utilize the GPU's peak compute capability. In contrast, VQVLA restructures VQ computation around spatial aggregation and temporal centroid reuse, and implements these reuse patterns with specialized hardware support, thereby achieving higher throughput.

Fig.~\ref{fig-exp-speedup} also validates the necessity of our VQVLA architecture, showing that GPU-VQVLA suffers from a $45.2\%$ performance loss compared to GPU-A100. Two fundamental limitations contribute to this performance degradation: 1) The spatial merging and temporal reusing techniques require traversing the index matrix and performing index comparisons, operations that GPUs handle inefficiently due to their limited capability for irregular logic processing. 2) The VQVLA algorithm requires multiple kernel invocations, including the robotic system state prediction, dynamic weight loading, spatial merging, and temporal reusing technique. Although CUDA supports concurrent kernel execution, we have observed that it is difficult to effectively overlap these decomposed kernels, as discussed in previous works\cite{zhao2021exploiting}.

\begin{figure}[h]
\centering
\includegraphics[width=0.9\linewidth]{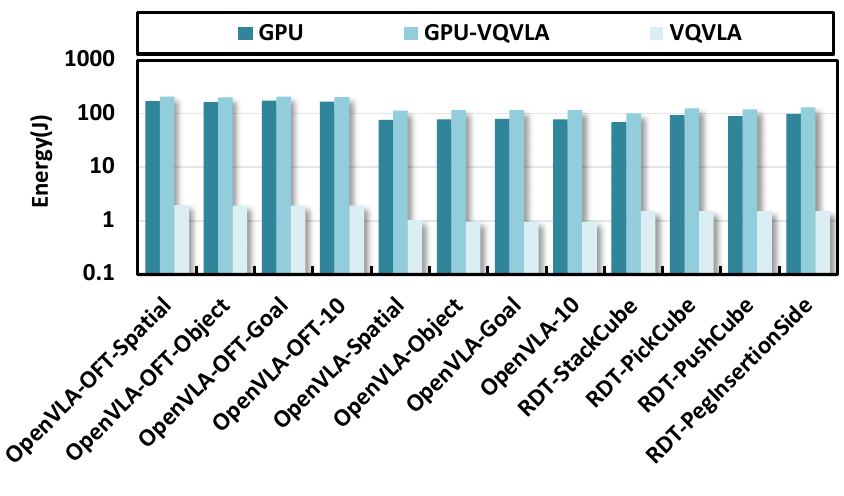}\vspace{-12pt}
\caption{Energy consumption of VQVLA and GPU.}\vspace{-12pt}
\label{fig-exp-energy}
\end{figure}

The energy results are depicted in Fig.~\ref{fig-exp-energy}. The VQVLA architecture delivers remarkable energy efficiency, surpassing GPU by $75.5\times$. These substantial savings in energy consumption come from three factors: (1) reduced off-chip memory accesses through adaptive fetching of the relevant codebooks and indices based on system states; (2) substantial elimination of redundant computations caused by repeated indices; and (3) the overall power consumption of VQVLA is only 6.4\% of that of the A100 GPU.

\begin{figure}[h]
\centering
\includegraphics[width=0.9\linewidth]{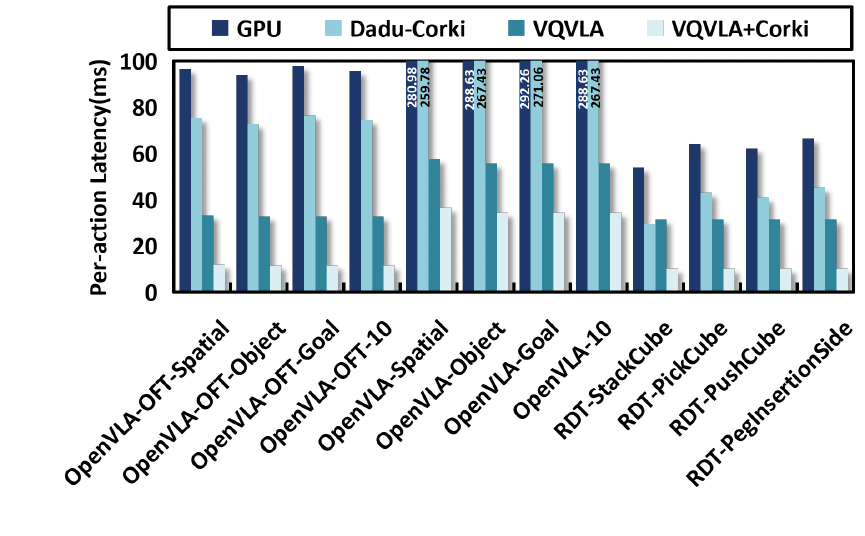}\vspace{-12pt}
\caption{The end-to-end per-action latency of embodied AI systems.}\vspace{-12pt}
\label{fig-exp-end-to-end}
\end{figure}

\textbf{End-to-end Per-Action Latency Improvement over Dadu-Corki.} To compare with Dadu-Corki, we evaluate the per-action latency in embodied AI systems, as shown in Fig.~\ref{fig-exp-end-to-end}. The per-action latency is computed by normalizing the end-to-end inference latency by the action chunk length. The end-to-end inference latency includes the time of VLA inference, robot control, and data communication, following the same methodology as Dadu-Corki. VQVLA achieves effective per-action latencies of 30-60ms across all evaluated models, corresponding to a $2.8\times$ performance improvement over Dadu-Corki. This advantage arises because VQVLA eliminates both computational and memory inefficiencies within VLA models, whereas Dadu-Corki primarily reduces the number of VLA inference calls and the latency of robot control. Furthermore, VQVLA can be integrated with Dadu-Corki (denoted as VQVLA-Corki), yielding a $6.0\times$ speedup compared to Dadu-Corki alone.

\begin{table}[h]
\centering
\caption{Area and power of the VQVLA Architecture.}
\vspace{-0.2cm}
\label{table-hardware}
\begin{tabular}{|cc|c|c|}
\hline
\multicolumn{1}{|c|}{\textbf{Module}}                                                                             & \textbf{Configuration}                                                                        & \textbf{\begin{tabular}[c]{@{}c@{}}Area\\ ($mm^2$)\end{tabular}} & \textbf{\begin{tabular}[c]{@{}c@{}}Power\\ ($W$)\end{tabular}} \\ \hline
\multicolumn{1}{|c|}{\begin{tabular}[c]{@{}c@{}}Index \\ Processing\\ Engine\end{tabular}}                        & \begin{tabular}[c]{@{}c@{}}Index Processing Units\\ (7MB SRAM)\end{tabular}              & 27.93                                                         & 9.00                                                         \\ \hline
\multicolumn{1}{|c|}{\multirow{6}{*}{\begin{tabular}[c]{@{}c@{}}Matrix \\ Multiplication\\  Engine\end{tabular}}} & \begin{tabular}[c]{@{}c@{}}Spatial Merging PE Array\\ (24×256 32-bit PEs)\end{tabular}        & 2.44                                                          & 0.69                                                         \\ \cline{2-4} 
\multicolumn{1}{|c|}{}                                                                                            & \begin{tabular}[c]{@{}c@{}}Temporal Reusing PE Array\\ (24×256 32-bit PEs)\end{tabular}       & 1.76                                                          & 0.50                                                         \\ \cline{2-4} 
\multicolumn{1}{|c|}{}                                                                                            & \begin{tabular}[c]{@{}c@{}}Adder Tree Unit\\ (24×128 3-stage Adder Trees)\end{tabular}        & 0.42                                                          & 0.31                                                         \\ \cline{2-4} 
\multicolumn{1}{|c|}{}                                                                                            & \begin{tabular}[c]{@{}c@{}}Result Cache\\ (1.5MB SRAM)\end{tabular}                           & 15.93                                                         & 7.97                                                         \\ \cline{2-4} 
\multicolumn{1}{|c|}{}                                                                                            & \begin{tabular}[c]{@{}c@{}}Cache Initialization Unit\\ (2048 32-bit Multipliers)\end{tabular} & 0.26                                                          & 0.04                                                         \\ \cline{2-4} 
\multicolumn{1}{|c|}{}                                                                                            & \begin{tabular}[c]{@{}c@{}}Accumulation Unit\\ (256 32-bit Adders)\end{tabular}               & 5.0e-3                                                        & 1.8e-4                                                       \\ \hline
\multicolumn{1}{|c|}{\begin{tabular}[c]{@{}c@{}}State \\ Predictor\end{tabular}}                                  & \begin{tabular}[c]{@{}c@{}}Distance Compute Unit\end{tabular}            & 2.4e-3                                                        & 3.5e-4                                                       \\ \hline
\multicolumn{1}{|c|}{\begin{tabular}[c]{@{}c@{}}On-chip \\ Buffer\end{tabular}}                                   & \begin{tabular}[c]{@{}c@{}}On-chip Buffer\\ (5.6MB SRAM)\end{tabular}                         & 2.40                                                          & 0.77                                                         \\ \hline
\multicolumn{2}{|c|}{\textbf{Total}}                                                                                                                                                                              & 51.15                                                         & 19.28                                                        \\ \hline
\end{tabular}
\end{table}

\textbf{Hardware Overhead and Area.} Table~\ref{table-hardware} provides a breakdown of configurations, area, and power of the VQVLA architecture. The matrix multiplication engine consists of two distinct PE arrays. The spatial merging PE array has $24\times 256$ PEs along with $24\times 128$ 3-stage adder trees; each PE integrates three multipliers and three adders. The temporal reusing PE array also consists of $24\times 256$ PEs and a 1.5MB result cache, with each PE containing two multipliers and three adders. The design rationale for the adder tree, result cache, and spatial-to-temporal PE ratio is explained in Section~\ref{ssect:exp-exploration}. The additional hardware area overhead from the IPU and result cache can be reduced for deployment in resource-constrained robotic systems, with only overflowed intermediate data spilled to off-chip memory. This provides a trade-off between on-chip resource usage and memory access latency.


\begin{figure}[h]
\centering
\includegraphics[width=0.8\linewidth]{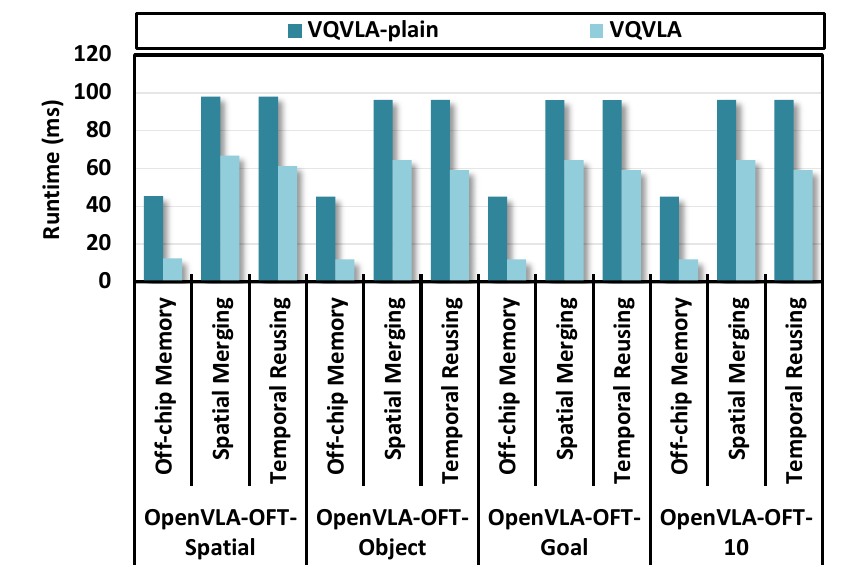}\vspace{-12pt}
\caption{Detailed analysis of contributions.}\vspace{-12pt}
\label{fig-exp-breakdown}
\end{figure}

\textbf{Ablation Study.} As shown in Fig.~\ref{fig-exp-breakdown}, we evaluate the contribution of key hardware components by comparing the proposed VQVLA with a baseline version (denoted as VQVLA-plain, which excludes all optimizations). All experiments are conducted on the OpenVLA-OFT model: 1. Off-chip memory: VQVLA selectively loads high- or low-precision sets according to the detected robotic system state, achieving a $3.8\times$ reduction in memory access latency. 2. Spatial merging PE array: VQVLA minimizes redundant computations within the same column by first accumulating the associated inputs, resulting in a $1.5\times$ latency reduction. 3. Temporal reusing PE array: VQVLA reduces repeated computations across columns by introducing a result cache to reuse previously computed results, leading to a $1.6\times$ latency reduction. 



\subsection{Design Exploration}\label{ssect:exp-exploration}

The performance of the VQVLA architecture is influenced by the capacity of the result cache, the number of stages in the adder tree, the PE ratio, and the threshold $T_d$. In this section, we determine the design parameters by balancing performance, task success rate, and hardware resource constraints. All experiments are conducted using three models: OpenVLA, OpenVLA-OFT, and RDT.

\begin{figure*}[!t]
\centering
\includegraphics[width=0.9\linewidth]{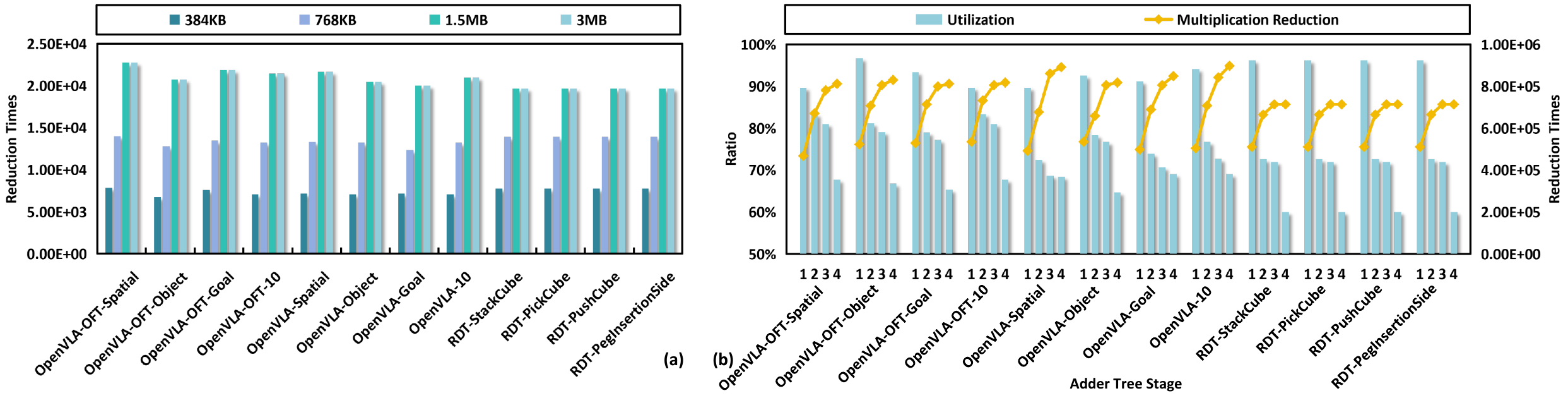}\vspace{-8pt}
\caption{Analyzation of the result cache and adder tree.}\vspace{-8pt}
\label{fig-exp-exploration-hardware}
\end{figure*}

\begin{figure}[!t]
\centering
\includegraphics[width=0.9\linewidth]{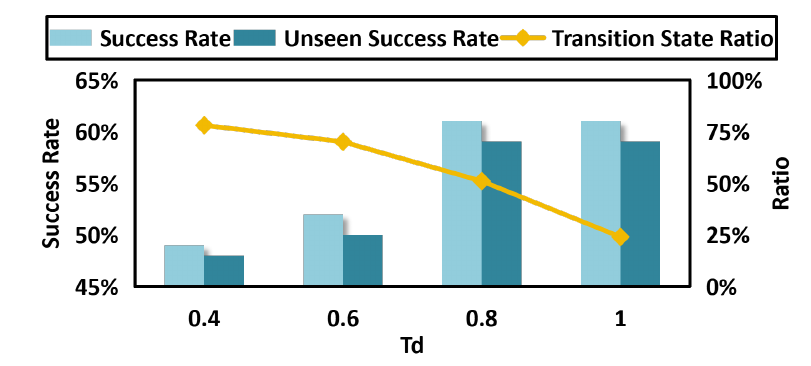}\vspace{-8pt}
\caption{Analyzation of the threshold $T_d$.}\vspace{-8pt}
\label{fig-exp-threshold}
\end{figure}

\textbf{Exploration of the capacity of the result cache.} A larger result cache enables more previously computed results to be retained, thereby increasing the likelihood of reuse during VLA inference. To identify the optimal capacity, we vary the cache size from 384KB to 3MB and evaluate the corresponding reduction in multiplications. As shown in Fig.~\ref{fig-exp-exploration-hardware}(a), for the OpenVLA-OFT model on the LIBERO-Spatial task, enlarging the cache to 1.5MB increases the number of reduced multiplications from 7k to 22k, leading to both improved performance and lower energy consumption. However, further increasing the cache size to 3MB yields no additional benefit, as a 1.5MB cache already captures all redundant computations. A similar trend is observed for OpenVLA on the LIBERO-Goal task, where the 1.5MB configuration also achieves near-saturated reuse. Therefore, we adopt 1.5MB as the configuration for all benchmarks.

\textbf{Exploration of the number of stages in the adder tree.} A larger number of stages in the adder tree enables the accumulation of more inputs per cycle, resulting in more multiplication reduction. However, excessively increasing the number of stages can lower the hardware utilization, as it becomes difficult to partition the spatial centroid location table into multiple sets without leaving adders idle. As illustrated in Fig.~\ref{fig-exp-exploration-hardware}(b), we evaluate the reduction in multiplications and the hardware utilization while varying the number of stages from 1 to 4. The results indicate that three stages achieve the best trade-off for all benchmarks, yielding the highest multiplication reduction and enough hardware utilization simultaneously.

\textbf{Exploration of the spatial-to-temporal PE ratio.} The spatial-to-temporal PE ratio determines how compute resources are allocated between the spatial and temporal PE arrays. A larger ratio favors exploiting centroid repetition across weight columns, whereas a smaller ratio allocates more resources to capturing centroid repetition within each weight column. As shown in Fig.~\ref{fig-exp-pe-utilization}, we evaluate three spatial-to-temporal PE ratios: 1:1, 1:2, and 2:1. The results show that the 1:1 ratio achieves the best performance across most benchmarks, as it provides a balanced resource allocation between the two reuse patterns and improves PE utilization to 90.6\%. Therefore, we adopt 1:1 as the spatial-to-temporal PE ratio.

\textbf{Exploration of the threshold $T_d$.} In VQVLA, the threshold $T_d$ controls the classification of system states into transition states, where low-precision weight sets can be fetched to reduce the memory footprint. A smaller $T_d$ classifies more states as transition states, leading to higher memory savings but potentially degrading the task success rate. To avoid overfitting to a specific workload, we randomly sample only 10\% of the data from each benchmark as the calibration set for selecting $T_d$, and evaluate the selected threshold on the remaining 90\% unseen data. To study its impact, we sweep $T_d$ from 0.4 to 1.0 on the calibration set. As shown in Fig.~\ref{fig-exp-threshold}, reducing $T_d$ from 1.0 to 0.8 increases the proportion of transition states, thereby improving memory savings. However, further reducing $T_d$ to 0.6 leads to a sharp decline in the success rate. Therefore, we set $T_d=0.8$ as the default threshold. When applying this calibrated threshold to the unseen test data, the success-rate degradation remains below 1.8\%, indicating that the selected threshold generalizes well across unseen scenarios.

%% file: Related.tex
\section{Related Works}

\subsection{Robotic System Optimization}
Traditional robots depend on optimization-based algorithms for decision-making and task planning~\cite{hao2024orianna,zhang2014loam}. These algorithms involve approximately solving linear equations to find the optimal solutions, which introduces considerable computational overhead but can only complete simple tasks. In contrast, emerging robotic applications~\cite{li2023vision,wake2024gpt,hu2023toward,firoozi2023foundation} use VLA models to control robots for tasks such as object manipulation, task planning, and navigation. These deep learning-based algorithms can complete complex tasks, demonstrating capabilities far superior to traditional algorithms.

As robots are increasingly treated as the next generation of computing platforms, the embodied AI community has increasingly focused on dedicated accelerators for robotic computing. Many accelerators have been designed for localization~\cite{liu2019eslam,suleiman2019navion,gan2021eudoxus,liu2021archytas,liu2022mobilesp,sugiura2022universal,eyvazpour2023hardware,sugiura2021unified}, motion planning~\cite{bakhshalipour2022racod,huang2024moped,hao2023blitzcrank,hsiao2023vapr,murray2019programmable,murray2016microarchitecture,lian2018dadu}, control~\cite{lian2017dadu, yang2023dadu,neuman2023roboshape,sacks2018robox,gac2012fpga,shao2018towards}, and navigation~\cite{yu2020building,krishnan2022automatic,lee2024spade}. However, these accelerators only focus on traditional optimization-based algorithms. In contrast, our work focuses on combining innovations in both algorithms and architecture to accelerate deep learning-based algorithms, distinguishing our work from previous research.

\subsection{Quantization Accelerator}
To achieve ultra-high execution performance in deep learning models, specialized quantization accelerators have emerged as a critical area of focus. These accelerators leverage low-bit quantization techniques~\cite{bai2020binarybert,zhou2025binary,ji2024beta,mao2020energy,ramachandran2025microscopiq} or exploit weight similarity post-quantization~\cite{li2025lut,parkcodegemm,jeon2020biqgemm,guo2024fast} to reduce computational overhead and memory usage.

Low-bit quantization methods, such as binary quantization~\cite{bai2020binarybert}, reduce computational requirements by using simple operations like XNOR gates~\cite{conti2018xnor}, but this approach often results in accuracy loss, especially when applied to complex models like VLA. As such, recent works have explored strategies that optimize the tradeoff between performance and accuracy. 
For example, MicroScopiQ~\cite{ramachandran2025microscopiq} is an outlier-aware microscaling quantization framework that preserves high-precision outliers while pruning low-importance weights to redistribute the extra outlier bits without breaking memory alignment. It further introduces a ReCoN-based accelerator to efficiently support mixed inlier/outlier computation, achieving high quantization accuracy with faster and more energy-efficient inference. These quantization accelerators~\cite{ramachandran2025microscopiq,hu2026m2xfp,chen2025p3,lee2025mx+} mainly improve performance through element-wise mixed precision and microscaling quantization. In contrast, our proposed VQVLA targets vector quantization, which represents a group of weights using shared codebook centroids rather than quantizing each weight individually, leading to different computation and data-reuse patterns.

Weight similarity post-quantization has been exploited to improve the balance between these factors, as seen in methods like LUT-DLA~\cite{li2025lut}. In this method, all-to-all multiplications between every (input, centroid) pair are precomputed and stored, reducing the need for repeated calculations during inference. However, this approach has limitations, such as inefficient use of memory for cold indices and redundant computations for rarely used pairs. The proposed VQVLA differentiates from LUT-DLA by leveraging the concept of hot-cold index patterns, where only frequently accessed (hot) indices are computed and stored on-demand. This results in more efficient memory usage and computational savings, ensuring that only necessary computations are performed when hot indices are encountered during inference.

%% file: Conclusion.tex
\section{Conclusion}

This paper presents VQVLA, an algorithm–hardware co-design framework for accelerating VLA inference. By analyzing the execution characteristics of robotic systems, we observe that VLA models exhibit state-dependent sensitivity and substantial centroid redundancy after VQ. Based on these insights, we propose MotionVQ, a motion-aware VQ scheme that dynamically adapts precision according to execution states, effectively balancing accuracy and memory efficiency. Furthermore, we introduce a merged-centroid vectorized GEMM paradigm that eliminates redundant computations by exploiting spatial and temporal locality of centroids. Extensive experiments show that VQVLA delivers satisfactory performance gain with acceptable success rate degradation.